\documentclass[11pt]{report}
\usepackage{rotating}
\usepackage{placeins}
\usepackage{geometry, graphicx, kotex, imakeidx, titlesec, array} 
\usepackage{mathptmx}
\usepackage{amsmath, amsthm, amssymb, mathrsfs, multirow, verbatim} 
\usepackage[nobiblatex]{xurl} 
\usepackage{indentfirst} 
\usepackage{booktabs} 
\usepackage{ragged2e}
\usepackage{setspace}
\titleformat{\chapter}{\normalfont\huge\bfseries}{\chaptertitlename\ \thechapter.}{20pt}{\huge} 
\makeindex 

\renewcommand\arraystretch{1.3} 

\begin{document}

\newpage   
\begin{center}
\pagenumbering{gobble} 
{\fontsize{16pt}{16pt}\selectfont Master's Thesis \par}  
\vspace{3cm} 
\huge Noise-Robust Abstractive Compression in Retrieval-Augmented Language Models
\par\vspace{4.5cm} 
{\fontsize{16pt}{16pt}\selectfont Singon Kim \par}  
\vspace{0.5cm}
{\fontsize{16pt}{16pt}\selectfont Department of Artificial Intelligence \par}  
\vspace{1.5cm}
{\fontsize{18pt}{18pt}\selectfont Graduate School \par} 
\vspace{0.5cm}
{\fontsize{18pt}{18pt}\selectfont Korea University \par} 
\vspace{1cm}
{\fontsize{14pt}{14pt}\selectfont December 2025} 
\end{center}

\newpage 
\begin{center}
\huge Noise-Robust Abstractive Compression in Retrieval-Augmented Language Models
\par\vspace{2.1cm} 
{\fontsize{16pt}{16pt}\selectfont by \par Singon Kim \par}
\vspace{0.5cm}
\rule{.6\textwidth}{0.4pt} 
\par\vspace{0.2cm}
{\fontsize{16pt}{16pt}\selectfont \mbox{under the supervision of Professor Seong-Whan Lee} \par}
\vspace{0.7cm}
{\fontsize{16pt}{18pt}\selectfont A thesis submitted in partial fulfillment of \par
 the requirements for the degree of \par Master of Science \par }   
\vspace{10pt}
{\fontsize{16pt}{16pt}\selectfont Department of Artificial Intelligence} 
\par\vspace{1.5cm}
{\fontsize{18pt}{18pt}\selectfont Graduate School \par \vspace{0.5cm}
 Korea University \par} 
\vspace{1cm}
{\fontsize{14pt}{14pt}\selectfont December 2025} 
\end{center}


\newpage 
\begin{center}
\vspace{1cm}
{\fontsize{16pt}{18pt}\selectfont
The thesis of Singon Kim has been approved \par
by the thesis committee in partial fulfillment\par
of the requirements for the degree of \par
Master of Science \par}  
\vspace{1cm}
{\fontsize{14pt}{14pt}\selectfont December 2025} 
\par\vspace{3cm}
\rule{.6\textwidth}{0.4pt}\par 
{\fontsize{16pt}{16pt}\selectfont Committee Chair: Seong-Whan Lee \par}
\vspace{1cm}
\rule{.6\textwidth}{0.4pt}\par 
{\fontsize{16pt}{16pt}\selectfont Committee Member: Won-Zoo Chung \par}
\vspace{1cm}
\rule{.6\textwidth}{0.4pt}\par 
{\fontsize{16pt}{16pt}\selectfont Committee Member: Tae-Eui Kam \par}
\end{center}

\newpage 
\pagenumbering{roman} 
\newgeometry{paper=b5paper, left=20mm, right=20mm, top=30mm, bottom=30mm} 
\addcontentsline{toc}{chapter}{Abstract}
\begin{center}
\LARGE Noise-Robust Abstractive Compression in Retrieval-Augmented Language Models 
\par\vspace{20pt}

\normalsize \doublespacing
by Singon Kim\par 
Department of Artificial Intelligence\par
under the supervision of Professor Seong-Whan Lee 
\par\vspace{20pt}
\large \textbf{Abstract}
\end{center}

\normalsize
\justifying 
\doublespacing

Abstractive compression utilizes smaller langauge models to condense query-relevant context, reducing computational costs in retrieval-augmented generation (RAG). However, retrieved documents often include information that is either irrelevant to answering the query or misleading due to factual incorrect content, despite having high relevance scores. This behavior indicates that abstractive compressors are more likely to omit important information essential for the correct answer, especially in long contexts where attention dispersion occurs. To address this issue, we categorize retrieved documents in a more fine-grained manner and propose Abstractive Compression Robust against Noise (ACoRN), which introduces two novel training steps. First, we use offline data augmentation on the training dataset to enhance compressor robustness against two distinct types of retrieval noise. Second, since the language model based compressor cannot fully utilize information from multiple retrieved documents and exhibits positional bias, we perform finetuning to generate summaries centered around key information that directly supports the correct answer. Our experiments demonstrate that T5-large, trained with ACoRN as a compressor, improves EM and F1 scores while preserving the answer string, which could serve as direct evidence. ACoRN excels on datasets with many accuracy-reducing documents, making it highly useful in real-world scenarios.

\par\vspace{20pt}


\newpage 
\begin{center}
\LARGE 검색 증강 언어 모델의 노이즈에 강건한 요약식 압축 
\par\vspace{20pt}
\normalsize 김 신 곤\par 
인 공 지 능 학 과\par 
지 도 교 수:  이 성 환
\par\vspace{20pt}
\addcontentsline{toc}{chapter}{국문초록}
\large \textbf{국문 초록}
\end{center}

\normalsize 

요약식 압축은 더 작은 언어 모델을 활용하여 질의 관련 컨텍스트를 응축함으로써, 검색 증강 생성(Retrieval-Augmented Generation, RAG)의 계산 비용을 절감합니다. 하지만 검색된 문서는 높은 관련성 점수에도 불구하고 질의에 대한 답변과 관련이 없거나, 부정확한 정보를 포함하여 오해를 유발하는 문서를 종종 포함합니다. 이러한 현상은 요약식 압축기가, 특히 어텐션 분산이 발생하는 긴 컨텍스트에서, 정답에 필수적인 중요 정보를 생략할 가능성이 더 높다는 것을 시사합니다. 이 문제를 해결하기 위해, 우리는 검색된 문서를 더 세분화하여 분류하고, 두 가지 새로운 학습 단계를 도입하는 ACoRN을 제안합니다.첫째, 우리는 학습 데이터셋에 오프라인 데이터 증강을 사용하여 두 가지 유형의 검색 노이즈에 대한 압축기의 강건성을 향상시킵니다. 둘째, 언어 모델 기반 압축기는 여러 검색 문서의 정보를 완전히 활용하지 못하고 위치 편향을 보이므로, 정답을 직접적으로 뒷받침하는 핵심 정보 중심으로 요약을 생성하도록 미세조정을 수행합니다. 우리의 실험은 ACoRN으로 학습된 T5-large 압축기가 직접적인 증거가 될 수 있는 답변 문자열을 보존하면서도 EM 및 F1 점수를 향상시킨다는 것을 보여줍니다. ACoRN은 정확도를 저해하는 문서가 많은 데이터셋에서 뛰어난 성능을 보여, 실제 시나리오에서 매우 유용합니다.

\par \vspace{20pt}
\textbf{주제어} : 노이즈 강건성, 요약식 압축, 검색 증강 언어 모델, 대규모 언어 모델




\newpage
\chapter*{Preface}
\addcontentsline{toc}{chapter}{Preface}
\normalsize
This dissertation is submitted for the degree of Master of Science in Artificial Intelligence at Korea University. This work was completed within the Department of Artificial Intelligence at Korea University, advised by Professor Seong-Whan Lee. Part of this work has been presented at the 2025 IEEE International Joint Conference on Neural Networks (IJCNN 2025). 

I was the lead investigator for the projects where I was responsible for all major areas of concept formation, data collection and analysis, as well as the majority of manuscript composition. Neither this, nor any substantially similar dissertation has been or is being submitted for any other degree, diploma, or other qualification at any other university.

\newpage
\chapter*{Acknowledgment}
\addcontentsline{toc}{chapter}{Acknowledgment}

This research was supported by the Institute of Information \& Communications Technology Planning \& Evaluation (IITP) grant, funded by the Korea government (MSIT) (No. RS-2019-II190079 (Artificial Intelligence Graduate School Program (Korea University)), No. RS-2024-00436857 (Information Technology Research Center (ITRC)), No. RS-2024-00457882 (AI Re- search Hub Project), and No. RS-2024-00336673 (AI Technology for Interactive Communication of Language Impaired Individuals).

\renewcommand*\contentsname{Contents}
\addcontentsline{toc}{chapter}{Table of Contents}
\tableofcontents

\bigskip


\listoftables
\addcontentsline{toc}{chapter}{List of Tables}


\listoffigures
\addcontentsline{toc}{chapter}{List of Figures}





\chapter{Introduction} \label{chap:intro}
\pagenumbering{arabic} 
Retrieval-augmented language models (RALMs) \cite{b1,b2,b3,b4} have strong capabilities in both academic research and industrial applications within the area of natural language processing (NLP). Retrieval-augmented generation (RAG) process in RALMs was designed to improve the performance of large language models (LLMs) \cite{b17,b36, b44} by integrating external knowledge. It can be achieved by simply appending supporting documents to the input, without the need to update the LLMs. While this approach helps bridge knowledge gaps in LLMs, it increases computational costs, particularly for transformer-based LLMs due to encoding substantially more tokens \cite{b20,b41}. With scaling, this burden becomes even more significant \cite{b39}. To mitigate these issues without compromising critical information, abstractive compression has been proposed \cite{b12}.

\begin{figure}[!t]
    \centering
    \includegraphics[width=\columnwidth]{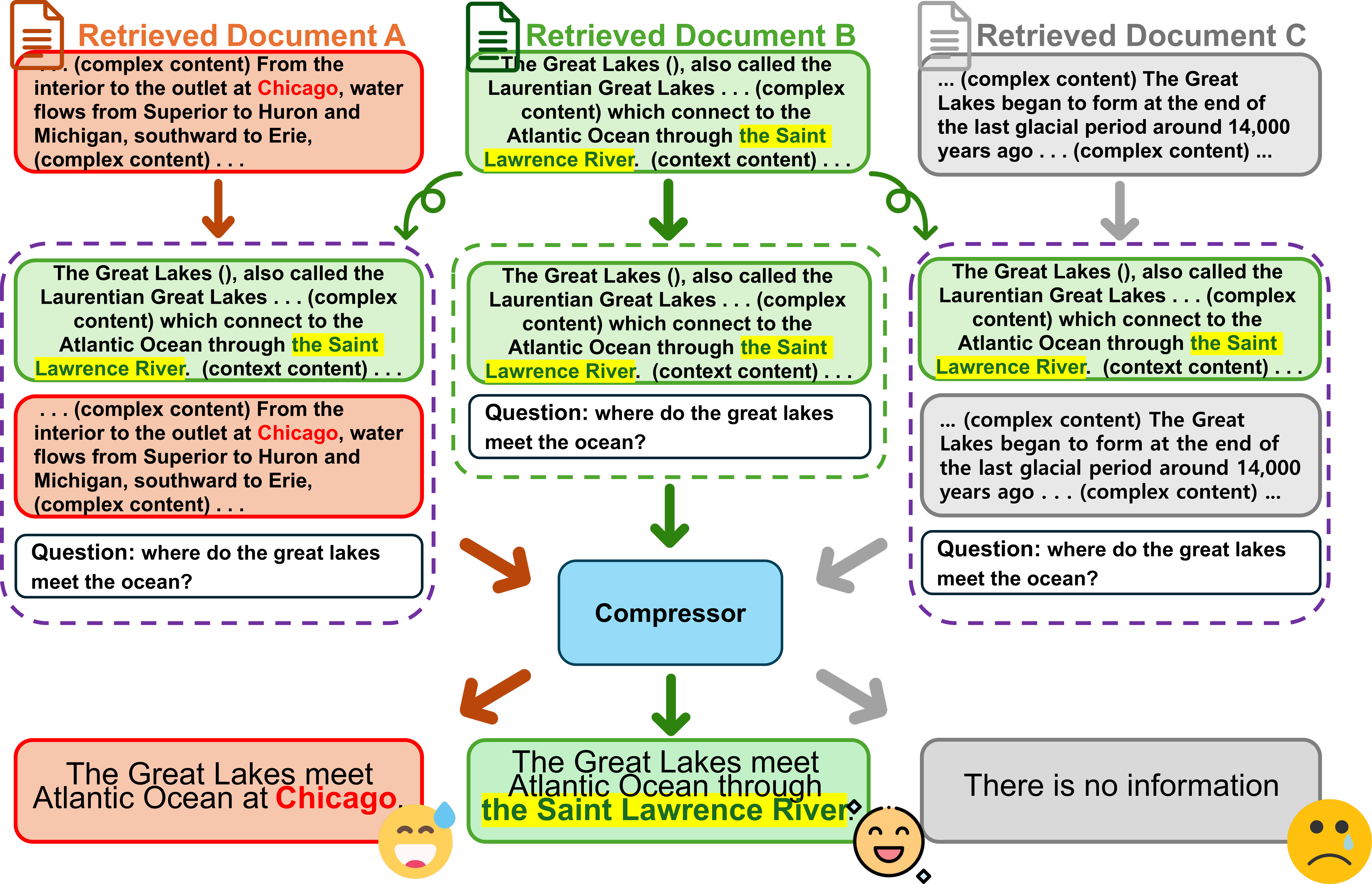}
    \caption{An illustrative example of a challenge in retrieving and summarizing information supporting to the correct answer from the documents. The compressor performs well in summarizing content supported to the correct answer when only the document including the correct answer is provided. However, it generates incorrect information or misses the key information when the retrieved documents contain inaccurate or irrelevant information.}
    \label{figure_1}
\end{figure}

Abstractive compression methods \cite{b12,b13} leverage the query-focused summarization (QFS) capabilities of language models \cite{b21} to reduce tokenization overhead effectively. However, existing abstractive compressors suffer from significant information loss when retrieving multiple documents. We suspect two issues based on previous studies. First, abstractive compressors are easily distracted not only by irrelevant information \cite{b8} but also by incorrect yet query-related information \cite{b25}. Second, in a long context, abstractive compressors do not utilize all available information for summarization \cite{b5,b34}. As shown in Fig. \ref{figure_1}, when only documents that directly support the correct answer are present, the compressor summarizes the content accurately. However, when irrelevant documents or documents providing incorrect answers are present alongside, it fails to summarize the relevant content properly. 

In this work, we seek to mitigate noise influence efficiently within the scope of the open-domain question answering (ODQA) \cite{b19} task. Then we propose Abstractive Compression Robust against Noise (\textit{ACoRN}), a training method that addresses this problem with the following two objectives: Mitigating distraction caused by information unrelated to correct answers, and reducing the loss of information that directly supports the correct answers in long contexts. 

Existing methods \cite{b12, b13} have not considered whether robust training on retrieval noise within the train dataset is sufficient to ensure robustness for the test dataset or real-world scenarios. However, building a training dataset that considers all previously proposed detailed analyses and classifications of retrieval noise \cite{b18, b25} can be challenging. Therefore, compared to previous approaches, we propose a concise noise classification specifically tailored for abstractive compression training: Retrieved documents that are thematically related to the query but contain incorrect information (\textit{Factual error documents}), and retrieved documents lacking sufficient information to answer the query (\textit{Irrelevant documents}). Additionally, factual error and irrelevant documents, collectively called \textit{noise documents}. In contrast to noise documents, we heuristically define \textit{Evidential documents} containing the answer string as positive documents regardless of whether they provide correct evidence to support the answer. Evidential documents are powerful to support accurate answer generation \cite{b27}. Detailed information can be found in Section \ref{Classifying Noise Documents}. Our method reconstructs the training dataset through offline data augmentation to ensure robustness against two types of retrieval noise, as described in Section \ref{Noise Documents Construction}.

To reduce the loss of information that directly supports the correct answer in long contexts, positional bias must be addressed. This problem, known as the ``lost in the middle'' phenomenon, often occurs in abstractive compression \cite{b5,b34}. To address this issue, inspired by FILM \cite{b40}, we propose fine-tuning that targets information directly supporting the correct answer. This approach preserves key information while mitigating positional bias by distributing evidential documents among noise documents within the compressor’s prompt. To generate labels that summarize information directly supporting the correct answer, we use data mining to extract and provide only evidential documents, instead of giving all retrieved documents to the LLM. This approach reduces the LLM's burden in handling long contexts and enhances its ability to focus on summarizing information that directly supports the correct answer.

Through these training steps, ACoRN enhances robustness against various types of retrieval noise and improves its ability to summarize key information within evidential documents. To demonstrate the effectiveness of our compression strategy for retrieving and summarizing evidential documents, we analyze the performance on ODQA benchmarks, including Natural Questions (NQ) \cite{b14}, TriviaQA \cite{b15}, and PopQA \cite{b6}. We also reconstruct benchmark test datasets by applying offline data augmentation to introduce various types of noise into the documents. Each test dataset is designed to demonstrate noise robustness from different perspectives. Our method, ACoRN, has shown improved performance over other compression methods. We also show that distilling LLM summarization abilities using evidential documents from top-\(k\) retrievals better preserves the answer string compared to using all top-\(k\) documents. Furthermore, in ACoRN we examine the impact of offline data augmentation on noise robustness.

The main contributions of our work can be summarized as follows:
\begin{itemize}
    \setlength{\itemsep}{0.5em} 
    \item A concise noise classification that distinguishes between irrelevant information and misleading yet query-related information, designed to enhance the robustness of abstractive compression training against retrieval noise.

    \item We show that extracting only evidential documents from the top-\(k\) retrieved results through data mining is an effective approach for generating summarization labels, compared to using the entire top-\(k\) results.

    \item We propose ACoRN, an effective and efficient noise-robust abstractive compression method. Validated on three ODQA benchmarks, it outperforms other methods, especially on datasets with a high noise-document ratio. 

\end{itemize}

\chapter{Related Works}\label{chap:related Works}

\section{Abstractive Compression} \label{subsec:Abstractive compression}

 Query-aware context compression methods summarize context differently depending on the question or task \cite{b23}. They can be categorized into token pruning \cite{b22, b33, b38, b45}, extractive compression \cite{b12, b24}, and abstractive compression \cite{b12,b13}. Unlike token pruning and extractive compression that simply focus on retaining the necessary tokens and sentences, abstractive compression retrieves and integrates only the essential information needed to answer the query and reformulates it accordingly. However, abstractive compression suffers from significant information loss, which is essential for supporting the correct answer to the query. This issue can be mainly attributed to positional bias, known as the ``lost in the middle'' phenomenon, where information in the middle of a given context is more likely to be omitted \cite{b5, b34, b46}.

Previous abstractive compression methods, such as COMPACT \cite{b13} and RECOMP \cite{b12}, have been proposed to overcome this drawback and enhance the effectiveness of compression. COMPACT \cite{b13} utilizes a large number of documents, dividing them into segment-size passages and iteratively retrieving the most important information to integrate it into summarization. This approach does not resolve the problem of information loss in abstractive compression, but rather makes up for it by retrieving more documents to summarize information in multiple steps. Despite its effectiveness, COMPACT requires higher computational cost and significantly increases the time required for compression.
RECOMP \cite{b12} performs data augmentation by generating multiple summarization samples to improve the quality of compressed passages. To train a model that focuses on summarizing key information about the query, the data with the lowest perplexity is selected among the samples that lead to generating the correct answer. However, generating summarization samples with API-driven LLMs imposes high costs, making it inefficient to construct an abstractive compression training dataset. Especially, when key information is absent or in intermediate positions, even generating many samples may lack quality. Unlike RECOMP, our method focuses on training the model to generate trustworthy labels curated for noise robustness without producing multiple samples.

\section{Noise-Robust Retrieval-Augmented Language Models} \label{subsec:Retrieval-Augmented Language Modeling with Noise Documents}

Retrieval-Augmented Language Models (RALMs) \cite{b1, b2, b3, b4, b42} refer to language models trained to generate answers based on documents retrieved from an external source. This approach has been demonstrated to improve model performance across a wide range of NLP tasks, including language modeling \cite{b35} and ODQA \cite{b19}. However, due to the limitations of the retriever's capabilities, retrieval-augmented systems inevitably have to deal with documents irrelevant or partially relevant to the task \cite{b10}. Prior studies \cite{b32, b37, b47} have shown that when the noise ratio in the retrieval context increases, the performance of RALMs noticeably declines \cite{b9}. 

One of previous noise robustness studies, RAAT \cite{b25}, has made two important observations. First, language models with fewer parameters are more easily distracted by noise documents. Second, in real-world scenarios there are various types of retrieval noise. It can be classified into three types: relevant retrieval noise, irrelevant retrieval noise, and counter factual retrieval noise. This demonstrates the need for training models against various types of retrieval noise. Additionally, Noise RAG Benchmark \cite{b18} defines as many as seven distinct noise types from a linguistic perspective. However, training robustness against various types of retrieval noise requires substantial computational cost, so such a fine-grained division can be challenging to implement. Thus, in contrast to previous approaches, we aim to propose a concise noise classification tailored specifically for abstractive compression training.

\chapter{Methodology}\label{chap:methodology}
\section{Overview} \label{subsec:overview}
\begin{sidewaysfigure}[!t] 
\centering
\includegraphics[width=\textheight]{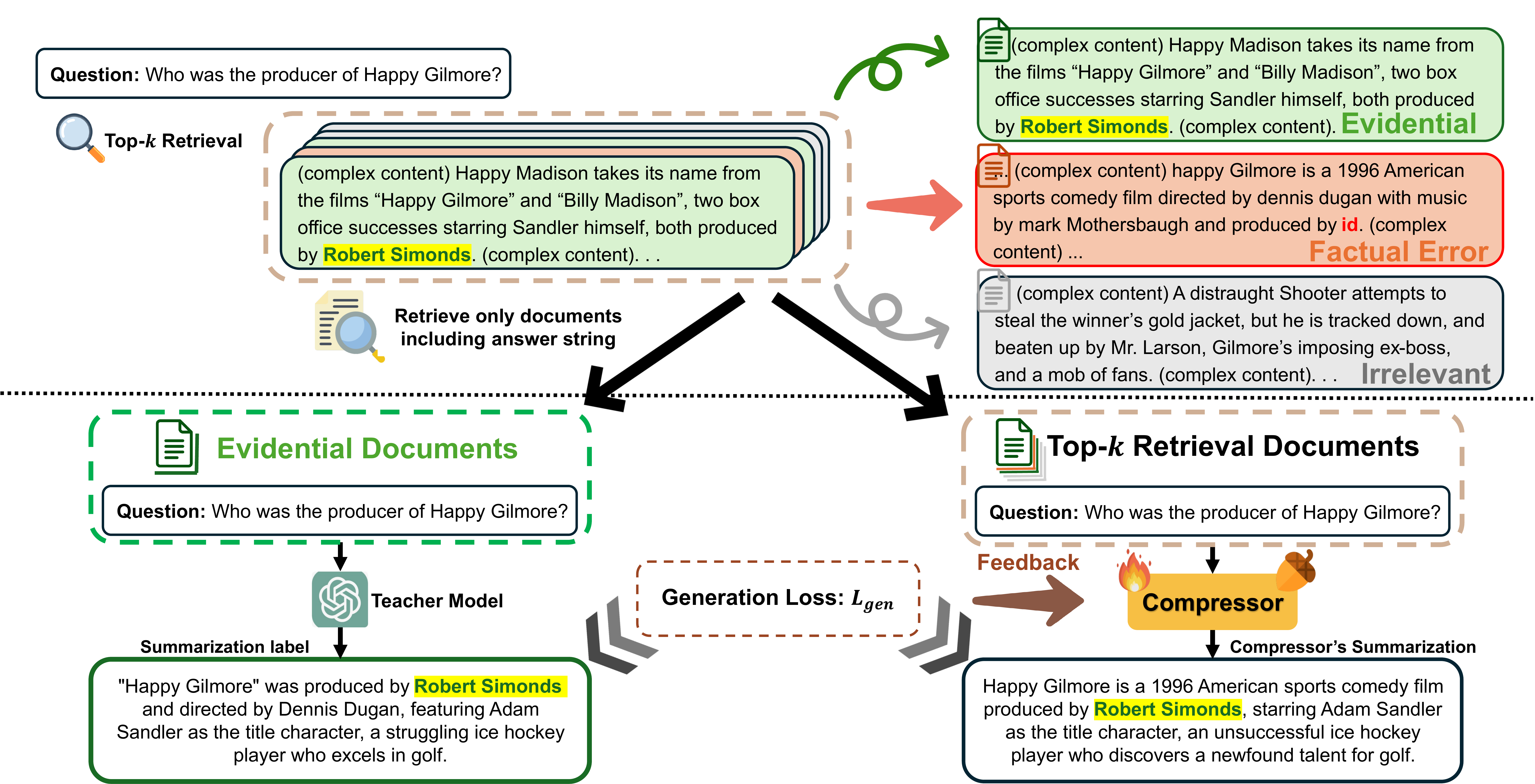} 
\caption{Overview of Abstractive Compression Robust against Noise (ACoRN). We fine-tune a compressor on our curated training dataset to make it robust against noisy documents and to retrieve evidential documents, focusing on summarizing the query based on the content of the evidential documents.}
\label{figure_2}
\end{sidewaysfigure}

We introduce Abstractive Compression Robust against Noise (ACoRN), shown in Fig. \ref{figure_2}. ACoRN is a simple, effective training approach enhancing robustness to noise from irrelevant retrievals and factual errors. The following subsections detail ACoRN.

In standard RALMs, given a query \( q \) is given, a retriever fetches top-\(k\) documents \( D_k = \{d_1, d_2, ... d_k\} \)  from an external database. A language model \( M \) predicts the answer \( y \) conditioned on \( q \), top-\(k\) documents \( D_k \), and an instruction \( I_{i} \). The instruction \( I_{i} \) guides \( M \) to generate \( y \) as follows: 
\begin{equation}
 M( I_{i}, D_k, q )=y .
\end{equation}

To reduce the computational cost of \( M \) caused by processing top-\(k\) retrieved documents, a compressor is introduced to summarize \(D_k\). Building on this approach, the goal can be formulated as follows: 
\begin{equation}
\arg\max_{\pi} P_M(y \mid I_{i}, S_\pi, q),
\end{equation}
\begin{equation}
S_\pi = \pi(I_{c}, D_k, q) \quad \text{with} \quad l(s_\pi) \ll l(D_k),
\end{equation}

\FloatBarrier

where \(\pi\) is a function that compresses documents \(D_k\) into a shorter context \(S_\pi\) based on the query \(q\), then \(l\) represents the number of tokens and \(I_c\) is the instruction to guide the compressor in summarizing the documents.
We divide \(D_k\) into two subsets: \(D_e\) and \(D_{noisy}\). \(D_e\) is the set of evidential documents, and \(D_{noisy}\) is the set of noise documents. If a retrieved document \(d\) contains the correct answer \(y\) about \(q\) we can denote \(d \in D_e\). However, if \(d\) doesn't contain \(y\), we denote \(d \in D_{noisy}\). Let \(S\) be the compressed context within the documents consisting of information that directly supports \(y\). 
We aim to fine-tune an abstractive compressor,  \(\pi^{'}\), that not only performs the mapping \( \pi^{'}: \{I_c, D_e, q\} \to S \), but also effectively summarizes important information supporting the correct answers, even in the presence of additional noise documents  \(D_{noisy}\). Formally, the function can be written as \( \pi^{'}: \{I_c, D_e, D_{noisy}, q\} \to S \).

\section{Classifying Noise Documents} \label{Classifying Noise Documents}

Existing studies \cite{b18, b25} on retrieval noise robustness classify types of noise commonly found in the real world. However, it is burdensome to consider all of them when training a compressor. Moreover, when using a detailed classification, some noise documents belong to multiple retrieval noise groups, further complicating the process. Therefore, we define the noise documents \(D_{noisy}\) to closely mimic real-world conditions but with minimal classification as follows:
\begin{equation}
D_{noisy} = D_{irr} \cup D_{f}.
\end{equation}

Here, \(D_{irr}\) represents the set of irrelevant documents. For any \(d \in D_{irr}\), this set encompasses contexts with high relevance to the \(q\), but where \(d\) lacks the information that directly supports \(y\). \(D_f\) is the set of factual error documents. For any \(d \in D_f\), this set includes contexts that are thematically related to \(q\) but contain incorrect or misleading information, such as incorrect historical facts or inaccurate numerical data. 

\begin{table}[!t]
\centering
\caption{Statistics for the three ODQA test datasets. \#Full represents total test data. \#Subset refers to the remaining data, controlled for noise type performance evaluation.}
\label{tab:my-table1}
\renewcommand{\arraystretch}{1} 
\setlength{\tabcolsep}{12pt} 
\begin{tabular}{cccc}
\hline
\multirow{2}{4em}{\textbf{Datasets}} & \multicolumn{3}{c}{\textbf{Test}} \\ 
\cline{2-4}
 & \textbf{\#Full} & \textbf{\#Subset} & \textbf{Percentage (\%)} \\ 
\hline
NQ \cite{b14}    & 3,610           & 1,417             & 39.25                    \\ 
TriviaQA \cite{b15}          & 11,313          & 2,966             & 26.21                    \\ 
PopQA \cite{b6}        & 1,399           & 413               & 29.52                    \\ 
\hline
\end{tabular}
\end{table}

To examine the influence of these distinct types of documents on compressors, we establish benchmarks for assessing retrieval noise robustness. Specifically, we established benchmarks for NQ \cite{b14}, TriviaQA \cite{b15}, and PopQA \cite{b6} individually, as detailed in Table \ref{tab:my-table1}. The details of the construction of this benchmark can be found in Section \ref{subsec:Datasets Construction}. Leveraging these benchmarks, we evaluate performance variations when the model is exposed only to evidential documents, compared to when irrelevant documents and factual error documents are additionally incorporated. As shown in Fig. \ref{figure_2}, we conduct experiments on Flan-T5-large \cite{b29}, analyzing the varying impacts of these three types of retrieved documents. The inclusion of evidential documents improves performance, whereas adding irrelevant or factual error documents results in a decline ranging from 5.68\% to 15.92\%. Through a comparative analysis of the effects of the two types of noise, we observe that the presence of irrelevant documents has minor impact on compressors with substantial capabilities.

\begin{figure}[!t]
    \centering
    \includegraphics[width=\columnwidth]{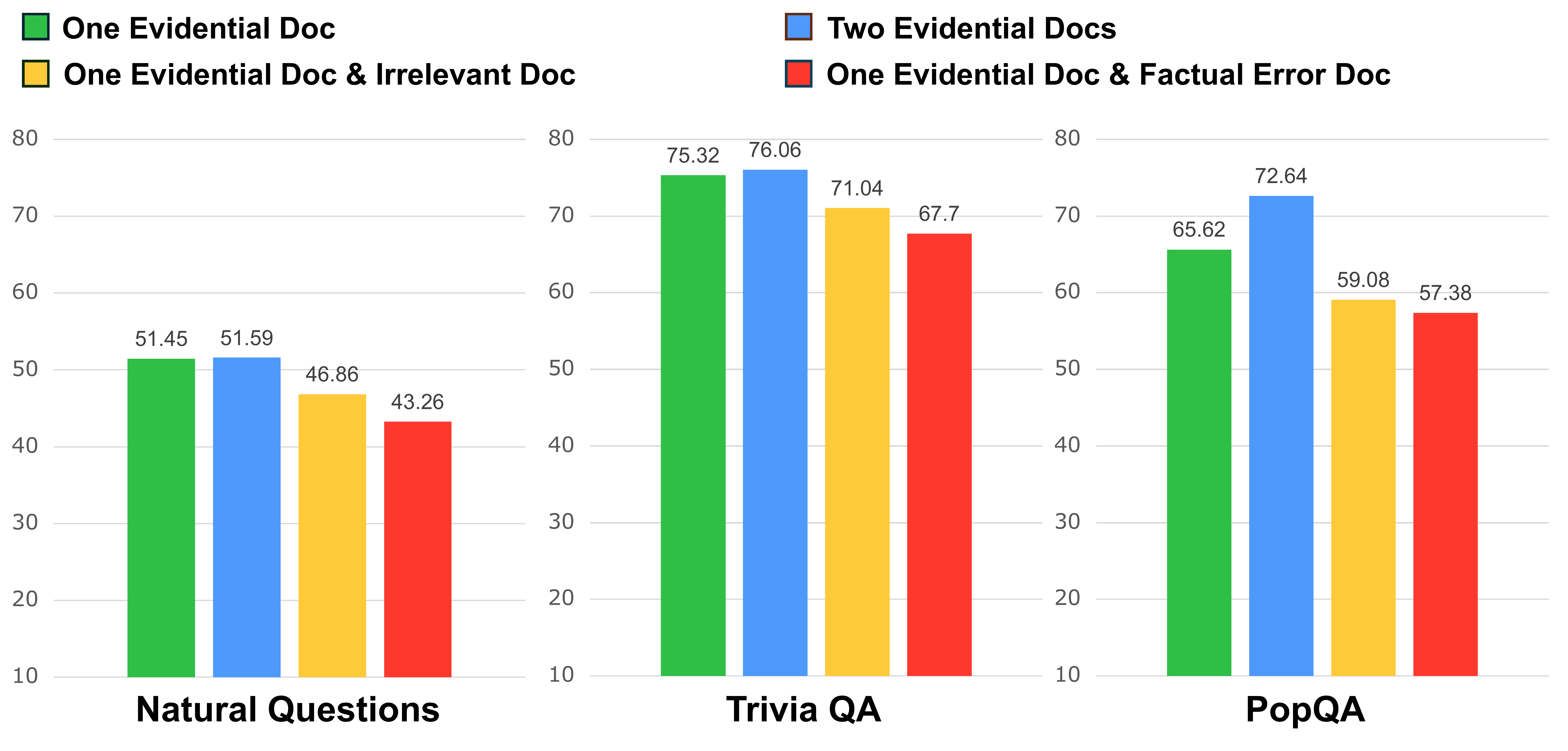}
    \caption{Exact Match (EM) scores for different types of noise documents, including irrelevant documents and factual error documents. Flan-T5-large \cite{b29} compresses documents using Query-Focused Summarization (QFS), compressed passages are then passed to LLaMA-3.1-8B-Instruct \cite{b36} to generate answers to the queries.}
    \label{figure_3}
\end{figure}

\section{Noise Documents Construction} \label{Noise Documents Construction}

Previous studies \cite{b32, b37} improved noise robustness by embedding retrieved noise documents within the fine-tuning data. Precisely calibrating both the kind and magnitude of noise is necessary to optimize the model’s performance \cite{b25}. To effectively incorporate the two types of retrieval noise in constructing our training dataset, we employ offline data augmentation \(D^{a}\). The main objective is to retrieve and summarize the information supporting \(y\) found in the evidential documents \(D_{e}^{a}\) even when noise documents are present.
According to Fig. \ref{figure_3}, factual error documents play a more critical role in model performance than irrelevant documents. For this reason, additional consideration is required for noise-robust training to address factual errors in documents.
We aim to train the compressor to recognize and summarize evidential documents when they provide conflicting information with factual error documents.
To ensure consistency, we do not augment factual error documents if evidential documents are absent. However, since evidential documents are not always present alongside factual error documents, we also intend to account for cases where evidential documents are present without any factual error documents during training. Hence, when there are retrieved \(N\) evidential documents, the probability of each evidential document becoming a factual error document is \(\frac {1}{N+1}\). Additionally, the probability of none of the evidential documents becoming factual error documents is also \(\frac {1}{N+1}\).
Formally, the set of evidential documents before offline data augmentation is \(D_e = \{e_1, e_2, . . . e_N\} \) and \(D_{ef}^{a}\) represents the collection of evidential documents that have been transformed into factual error documents through data augmentation. For an arbitrary number \(m \in \{1, 2, . . . N\} \) given, 

\begin{equation}
\begin{split}
    P(e_m \in D_{ef}^{a}) &= \frac {1}{N+1}, \\
    \text{with} \quad P\left(\bigcup_{j=1}^{N} e_j \in D_{ef}^{a} \right) &= \sum_{j=1}^{N} P(e_j \in D_{ef}^{a}).
\end{split}
\end{equation}

\section{Evidential Pseudo Labels for Summarization}

We train an abstractive compressor, taking an input sequence \(x = \{I_c, q\}\) along with a concatenation of top-\(k\) retrieved documents \(D_k\) and producing a summary \(S_\pi\). Since we desire the compressor to be substantially smaller than the large-scale model \(M\), we employ knowledge distillation \cite{b16} to build a lightweight abstractive compressor. For generating pseudo summarization labels \(S\) based on evidential documents, we use a pre-trained LLM as a teacher model. 
The teacher model is provided with a query and only the evidential documents to perform QFS via an instruction. In other words, the teacher model generates the pseudo summarization labels \(S\) by focusing on the evidential documents after offline data augmentation \( D_{e}^{a} \). Formally, the objective \( \pi_{t}\) of generating ground truth summarization labels is as follows:
\begin{equation}
S = \pi_{t}(I_{c}, D_{e}^{a}, q).
\end{equation}

If \( D_{e}^{a} = \emptyset \), we don't pass any retrieved documents to \(M\). Then \(M\) generates the answer without any supporting information as follows:
\begin{equation}
 M(I_{i}, q ) = y.
\end{equation}

\section{Abstractive Compression Training}

Using the training dataset constructed with \(S\) based on evidential documents, the compressor is distilled \cite{b16}. At this stage, the input differs from that of the teacher model as it includes noise documents  \( D_{noisy}^{a} \) such as  \( D_{irr}^{a} \) and  \( D_{f}^{a} \). To effectively train the compressor to summarize information based on evidential documents, a function is defined as follows:
\begin{equation}
S_\pi = \pi^{'}(I_c, D^a, q).
\end{equation}

The loss function is designed to facilitate summarization training by enforcing a strong alignment between the summarization labels \(S\) and the generated summaries \(S_\pi\).
In this setup, \(N\) denotes the sequence length in each sample. 
\(\theta\) denotes the parameters of the compressor. Then the loss function is expressed as:
\begin{equation}
L_{\text{gen}}(\theta, x, S) = -\sum_{i=1}^{N} \log P_{\theta}(S_{i} \mid x, S_{<i}).
\end{equation}

\chapter{Experiments}\label{chap:experiments}
\section{Implementation Details}
\subsection{Baselines}
We evaluate our approach in language models using ODQA benchmarks, specifically the NQ \cite{b14, b43}, TriviaQA \cite{b15} and PopQA \cite{b6} datasets. NQ comprises queries \(q\) paired with short answers containing no more than five tokens. TriviaQA is constructed by extracting spans from Wikipedia articles that contain correct answers to each given query. PopQA is created by transforming Wikidata knowledge triples, consisting of a subject, a relationship, and an object into natural language query using manually written templates for 16 diverse relationship types. For all experiments, documents are retrieved from Wikipedia using the adversarial Dense Passage Retriever (DPR) \cite{b31}, which finds five documents per query. For inference, we use Llama-3.1-8B-Instruct \cite{b36} as the language model \(M\) and perform our experiments in a zero-shot manner. As the compressor, Flan-T5-large \cite{b29} and T5-large \cite{b30} are fine-tuned for the task. We use two NVIDIA RTX A6000 GPUs for fine-tuning and a single NVIDIA RTX A5000 GPU for inference.
\subsection{Datasets Construction} \label{subsec:Datasets Construction}
To make the compressor resistant to retrieval noise, we create a training dataset and two test benchmarks using offline data augmentation. In the cases where two or more evidential documents are present in the top-\(5\) retrieved documents, we randomly select one or none of them, mask the answer entity and use RoBERTa-large \cite{b28} to replace it with an incorrect entity, resulting in a document with factual errors.

The training dataset \(T = \{q, D^a, S\}\) includes each query \(q\), offline-augmented documents \(D^a\), and summarization labels \(S\). To create more accurate pseudo labels \(S\), we use only the set of evidential documents \(D_e^a \subseteq D^a\) as supporting documents. As with a previous method \cite{b12}, we utilize the QFS capabilities of GPT-3.5-turbo. It is provided with \{\(q\), \(D_e^a\)\} to generate high quality QFS, which is then used as \(S\).

We design benchmarks to evaluate two aspects of noise robustness: 
\begin{itemize}
    \setlength{\itemsep}{0.5em} 
    \item \label{A1} Assessing whether the top-\(5\) retrieved documents, despite retrieval noise, can detect and compress the evidence directly supporting the correct answer, as discussed in Section \ref{results:3}.

    \item \label{B1} Evaluating the impact of incorporating various types of retrieved noise documents along with the evidential ones on compression quality, as described in Section \ref{results:4}.
\end{itemize}

 To evaluate the first one, we apply filtering to the queries, ensuring that each query in the filtered subset contains at least one evidential document from the offline data augmented test dataset. To assess the last one, we create a new dataset by extracting samples that include all types of retrieved documents: evidential documents, irrelevant documents, and factual error documents. Performance is then compared under these scenarios: An evidential document only, an evidential document combined with an irrelevant document, and an evidential document combined with a factual error document. The statistics of this newly formulated benchmark compared to the original datasets is shown in Table \ref{tab:my-table1}.

\subsection{Training}

We train the compressor using the training dataset \(T\), where \(D^a\) and \(q\) are provided as inputs with a compression instruction \(I_c\). The training objective is formally defined by the function \( \pi: \{I_c, D^a, q\} \to S \), where \(S\) represents the summarization labels produced by GPT-3.5-turbo, and the compressor is trained to replicate these labels, effectively distilling GPT-3.5-turbo's summarization capability. We conduct experiments with two models: Flan-T5-large and T5-large. We also provide instructions \(I_c\) only to the Flan-T5-large. The batch size per device is set to 2, with gradient accumulation step of 2. Evaluation is performed every 1000 steps.

\subsection{Evaluation Metrics}

We assess the performance of our method by measuring the Exact Match (EM) score and F1 score, and we also evaluate efficiency using the compression ratio (CR) \cite{b7} and inference time \cite{b11}. Specifically, EM measures how precisely the system's answer matches the reference answer at the character level, while the F1 score balances precision and recall, evaluating the accuracy of identified answers and minimizing omissions. CR evaluates how efficiently the compressor summarizes the information essential to answer the query. Inference time refers to the time taken by the language model \(M\) to process and generate a response for input data during inference. Inference time is critical for practical applications requiring real-time user query responses or large-scale data processing. The unit of inference time is seconds.

\section{Results}


\begin{sidewaystable} 
\centering
\caption{Quantitative evaluation of ACoRN on the ODQA tasks using LLaMA-3.1-8B-Instruct}
\label{tab:my-table2}
\renewcommand{\arraystretch}{1.5} 

\resizebox{\textheight}{!}{%
\begin{tabular}{lcccccccccccc}
\hline

\multirow{2}{4em}{\textbf{Method}} & \multicolumn{4}{c}{NQ} & \multicolumn{4}{c}{TriviaQA} & \multicolumn{4}{c}{PopQA} \\ 
\cline{2-13}

& \textbf{EM (↑)} & \textbf{F1(↑)} & \textbf{CR(↓)} & \textbf{Inference Time(↓)} 
& \textbf{EM (↑)} & \textbf{F1(↑)} & \textbf{CR(↓)} & \textbf{Inference Time(↓)} 
& \textbf{EM (↑)} & \textbf{F1(↑)} & \textbf{CR(↓)} & \textbf{Inference Time(↓)} \\
\hline

No Retrieval & 19.94 & 31.46 & - & 0.249 & 49.65 & 59.59 & - & 0.162 & 20.51 & 23.68 & - & 0.132 \\
Top-1 document & 27.87 & 42.12 & - & 0.241 & 48.21 & 61.67 & - & 0.216 & 17.87 & 21.01 & - & 0.154 \\
Top-5 documents & 31.83 & 47.40 & - & 0.444 & 57.90 & 69.38 & - & 0.246 & 41.46 & 51.02 & - & 0.321 \\
\hline
LongLLMLingua \cite{b22} & 26.84 & 42.72 & 0.562 & 0.396 & 54.45 & 66.59 & 0.562 & 0.297 & 38.24 & 47.87 & 0.597 & 0.238 \\
Quito \cite{b38} & 29.94 & 44.04 & 0.484 & 0.316 & 56.61 & 67.61 & 0.484 & 0.258 & 41.89 & 50.36 & 0.523 & 0.241 \\
RECOMP \cite{b12} & 32.58 & 45.42 & \textbf{0.052} & 0.225 & \textbf{58.64} & 68.40 & 0.049 & 0.174 & - & - & - & - \\
\textbf{ACoRN (T5-large)} & 34.97 & 47.70 & 0.064 & \textbf{0.205} & 58.20 & 68.33 & 0.049 & 0.174 & 45.60 & 52.38 & \textbf{0.059} & \textbf{0.137} \\
\hline
Flan-T5-large \cite{b29} & 31.97 & 45.15 & 0.133 & 0.227 & 57.34 & 67.72 & 0.085 & 0.183 & 42.89 & 48.80 & 0.071 & 0.142 \\
\textbf{ACoRN (Flan-T5-large)}& \textbf{35.56} & \textbf{48.48} & 0.065 & 0.208 & 58.33 & \textbf{68.58} & 0.056 & 0.174 & \textbf{45.75} & \textbf{52.82} & 0.062 & 0.139 \\
\hline
\end{tabular}%
}
\end{sidewaystable}
\subsection{Main Results} \label{Main Results}

Table \ref{tab:my-table2} presents our main results and illustrates the effectiveness of our method ACoRN, compared to the baselines in terms of EM, F1, CR, and inference time. First, we compared the results of our method to those from top-\(5\) retrieval without compression. Our experiments demonstrate that our method, ACoRN, minimizes the loss of critical information, effectively conveys it to the language model \(M\), and reduces inference time, resulting in higher EM scores across all datasets. This indicates that through ACoRN irrelevant documents and factual error information are eliminated, allowing \(M\) to confidently generate accurate answers. Second, we compared our method's results to those of other compression methods. Our approach outperforms existing token pruning methods such as LongLLMLingua \cite{b22} and Quito \cite{b38} across all metrics. Our method shows slightly lower metrics on TriviaQA compared to RECOMP \cite{b12}, which is attributed to the inherent differences of language model \(M\). As explained in Section \ref{subsec:Abstractive compression}, RECOMP uses perplexity during label creation to selectively augment data, optimizing summarization to improve the model's ability to answer correctly. Combining our method with RECOMP's selective data augmentation could yield even better results. Third, the summarization generated by ACoRN significantly reduces the latency of answer generation by the language model \(M\), as demonstrated by inference time. With analysis of EM, F1 scores, CR and inference time, we observe that by summarizing key information critical for answer generation, \(M\) was able to easily retrieve relevant information and generate accurate answers.

\subsection{Reliability of Summarization Labels} \label{Reliability of Summarization Labels}
\begin{figure}[!t]
    \centering
    \includegraphics[width=\columnwidth]
    {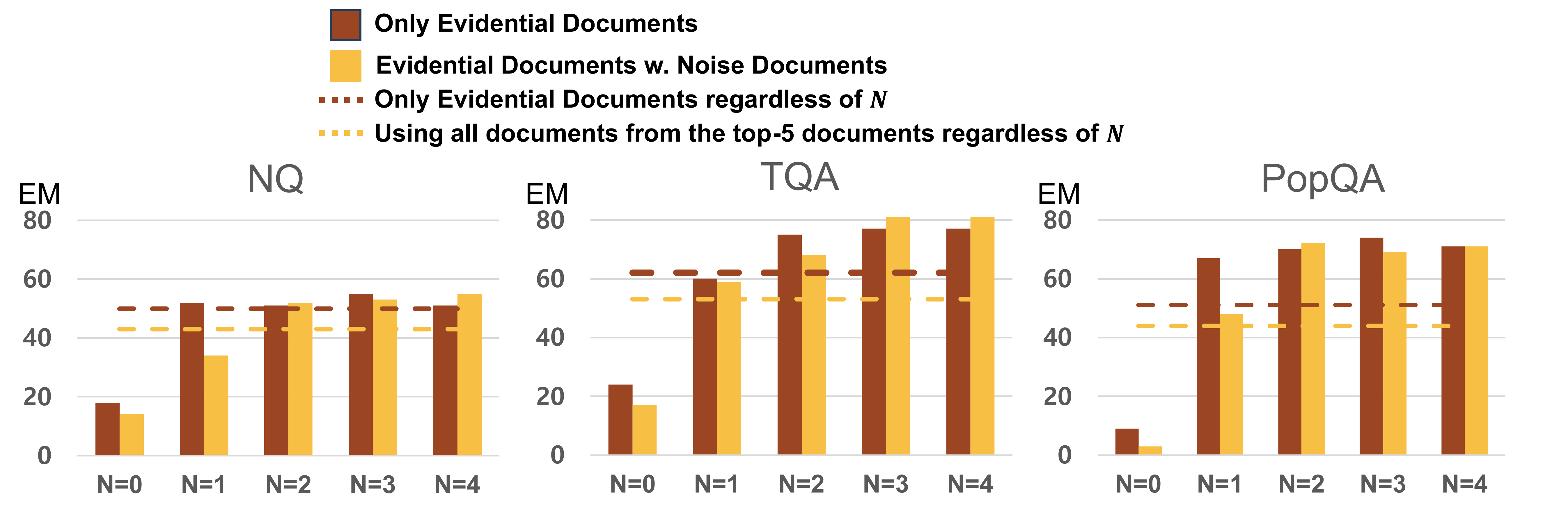}
    \caption{Performance comparison of GPT-3.5-turbo QFS using only evidential documents versus all top-5 documents in the prompt, based on 100 random samples for each evidential document count \(N\) in top-5 retrieval. The \(N\)=0 case (evidential only) means using only internal knowledge. The compressed output is passed to the inference model's \(M\) prompt, LLaMA-3.1-8B-Instruct. The dotted line represents performance from 100 randomly sampled instances, regardless of \(N\).}
    \label{figure_4}
\end{figure}
As shown in Fig. \ref{figure_4}, we compared the reliability of summarization labels generated using all documents versus using only evidential documents. We split the test dataset based on the number of evidential documents \(N\) from the top-\(5\) retrieved documents. We aim to examine how the quality of summarization labels changes with the proportion of presented evidential documents and how the performance gap between the two settings varies accordingly. We randomly sampled 100 instances for each \(N\) to ensure fair evaluation. We observe the summarization performance of the teacher model in two scenarios: when only \(N\) evidential documents are provided, and when \(N\) evidential documents are combined with \( 5 - N \) noise documents. When the fixed value of \(N\) is small, using only the evidential documents for summarization is more effective compared to using all retrieved documents. When we increase \(N\), there are cases where utilizing all retrieved documents is more effective. This is because the retrieval noise from documents with low semantic relevance to the query can help better differentiate evidential documents from noise, leading to a more effective summarization \cite{b18}. However, the effect is minimal when compared to using only evidential documents. Results show that using only evidential documents for summarization labels is more effective.

\subsection{Evaluation of Answer String Preservation} \label{results:3}

\begin{table}[]
\centering
\caption{Evaluation of Preserving Answer string Ratio (PAR) and compression ratio (CR) in offline augmented test dataset}
\label{tab:my-table3}
\resizebox{\columnwidth}{!}{%
\begin{tabular}{@{}lcccccc@{}} 
\toprule

\multirow{2}{4em}{\textbf{Method}} & \multicolumn{2}{c}{\textbf{NQ}} & \multicolumn{2}{c}{\textbf{TriviaQA}} & \multicolumn{2}{c}{\textbf{PopQA}} \\ 
\cmidrule(lr){2-7} 

& \textbf{CR(↓)} & \textbf{PAR(↑)} & \textbf{CR(↓)} & \textbf{PAR(↑)} & \textbf{CR(↓)} & \textbf{PAR(↑)} \\ 
\midrule

\textit{Token pruning} \\
\quad LongLLMLingua & 0.5607 & 0.6320 & 0.5625 & 0.7044 & 0.5963 & 0.7760 \\ 
\midrule 

\textit{Abstractive compression} \\
\quad RECOMP & \textbf{0.0557} & 0.4966 & 0.0522 & 0.6283 & - & - \\
\quad ACoRN (T5-large) & 0.0637 & 0.6425 & 0.0522 & 0.6380 & 0.0675 & 0.7521 \\
\quad ACoRN (Flan-T5-large) & 0.0663 & \textbf{0.6687} & 0.0579 & \textbf{0.6686} & \textbf{0.0674} & \textbf{0.7635} \\ 
\bottomrule

\end{tabular}%
}
\end{table}

When at least one evidential document exists for a query, we evaluate the preservation ratio of the answer string (PAR) in the compression output to assess how well the model summarizes content that directly supports the correct answer within the retrieved documents. As shown in Table \ref{tab:my-table3}, our method demonstrates a high compression ratio while maintaining a remarkably high PAR. Notably, when using T5-large, we increase PAR on NQ from \(0.4966\) to \(0.6425\) compared to RECOMP. We also achieve a PAR similar to LongLLMLingua, which uses the token pruning method. Since token pruning preserves key information based on a binary classification, it is inherently easier to preserve tokens compared to abstractive compression. This demonstrates that ACoRN successfully achieves its goal of recognizing and compressing evidential information from the retrieved documents.

\subsection{Impact of Training with Noise Documents Augmentation} \label{results:4}

\begin{table*}[]
\centering
\caption{Quantitative evaluation of ACoRN performance changes with increased retrieval noise through data augmentation in ODQA test dataset retrieved documents, compared to the method without noise documents augmentation in the training dataset.}
\label{tab:my-table4}
\resizebox{\textwidth}{!}{%
\begin{tabular}{@{}lcccccc@{}} 
\toprule

\multirow{2}{*}{Method} & \multicolumn{2}{c}{NQ} & \multicolumn{2}{c}{TriviaQA} & \multicolumn{2}{c}{PopQA} \\ 
\cmidrule(lr){2-7} 

& EM (↑) & F1 (↑) & EM (↑) & F1 (↑) & EM (↑) & F1 (↑) \\ 
\midrule

ACoRN (T5-large) 
& \begin{tabular}[c]{@{}c@{}}34.79 \(\to\) \textbf{32.68}\\ \textbf{(-2.11)}\end{tabular} 
& \begin{tabular}[c]{@{}c@{}}\textbf{47.70} \(\to\) \textbf{45.59}\\ \textbf{(-2.11)}\end{tabular} 
& \begin{tabular}[c]{@{}c@{}}58.20 \(\to\) 57.31 \\ \textbf{(-0.89)}\end{tabular} 
& \begin{tabular}[c]{@{}c@{}}68.33 \(\to\) \textbf{67.35}\\ \textbf{(-0.98)}\end{tabular} 
& \begin{tabular}[c]{@{}c@{}}\textbf{45.60} \(\to\) \textbf{44.39}\\ \textbf{(-1.21)}\end{tabular} 
& \begin{tabular}[c]{@{}c@{}}\textbf{52.38} \(\to\) \textbf{50.67}\\ \textbf{(-1.71)}\end{tabular} \\

\quad w/o noise documents augmentation 
& \begin{tabular}[c]{@{}c@{}}\textbf{35.15} \(\to\) 32.54\\ (-2.61)\end{tabular} 
& \begin{tabular}[c]{@{}c@{}}47.65 \(\to\) 44.88\\ (-2.77)\end{tabular} 
& \begin{tabular}[c]{@{}c@{}}\textbf{58.57} \(\to\) \textbf{57.38}\\ (-1.19)\end{tabular} 
& \begin{tabular}[c]{@{}c@{}}\textbf{68.54} \(\to\) 67.23\\ (-1.31)\end{tabular} 
& \begin{tabular}[c]{@{}c@{}}44.10 \(\to\) 41.82\\ (-2.28)\end{tabular} 
& \begin{tabular}[c]{@{}c@{}}50.37 \(\to\) 47.88\\ (-2.49)\end{tabular} \\

\bottomrule
\end{tabular}%
}
\end{table*}

\begin{sidewaystable} 
\centering
\caption{Performance comparison of training a compressor, built on T5-large, using factual error documents constructed through offline data augmentation versus using original retrieval documents on LLAMA-3.1-8B-Instruct}
\label{tab:my-table5}
\renewcommand{\arraystretch}{1.5} 

\resizebox{\textheight}{!}{%
\begin{tabular}{lcccccccccccc}
\hline

\multirow{2}{4em}{\textbf{Method}} & \multicolumn{2}{c}{One Evidential Doc} 
& \multicolumn{2}{c}{\begin{tabular}[c]{@{}c@{}}One Evidential Doc \\ w. Irrelevant Doc\end{tabular}} 
& \multicolumn{2}{c}{\begin{tabular}[c]{@{}c@{}}One Evidential Doc \\ w. Factual Error Doc\end{tabular}} \\ 
\cmidrule(lr){2-3} \cmidrule(lr){4-5} \cmidrule(lr){6-7} 

& EM (↑) & F1(↑) & EM (↑) & F1(↑) & EM (↑) & F1(↑) \\ 
\midrule

\textit{Natural Questions} \\
\quad ACoRN & \textbf{51.94} & 66.55 & \textbf{50.11} & 64.39 & \textbf{49.12} & \textbf{62.96} \\
\quad \quad w/o noise documents augmentation & 51.66 & \textbf{66.70} & 49.82 & \textbf{64.54} & 47.71 & 62.13 \\ 
\midrule

\textit{TriviaQA} \\
\quad ACoRN & 74.68 & 82.98 & \textbf{72.96} & \textbf{81.62} & \textbf{70.20} & \textbf{78.73} \\
\quad \quad w/o noise documents augmentation & \textbf{74.88} & \textbf{83.14} & 71.94 & 80.27 & 68.84 & 77.27 \\ 
\midrule

\textit{PopQA} \\
\quad ACoRN & \textbf{66.34} & \textbf{72.50} & \textbf{58.84} & \textbf{64.29} & \textbf{54.00} & \textbf{60.33} \\
\quad \quad w/o noise documents augmentation & 64.41 & 70.74 & 58.60 & 63.62 & 53.75 & 59.55 \\ 
\bottomrule
\hline
\end{tabular}%
}
\end{sidewaystable}

To gain an understanding of the individual contribution of each retrieval noise type within ACoRN to the overall performance, we conduct an ablation study by comparing the process with and without noise documents augmentation to train a compressor. First, we evaluated the performance of the two processes on three existing ODQA test datasets. Next, we reconstructed the ODQA test datasets by retrieved documents augmentation and compared the performance on these reconstructed datasets. Then, we analyzed the degree of performance degradation between the two test datasets. The noise documents augmentation process for training reduced the extent of performance degradation as the level of noise increased. The results are shown in Table \ref{tab:my-table4}.
Second, we compared the changes in EM and F1 scores when each retrieval noise type was incorporated to the evidential documents. When only evidential documents were present, the noise documents augmentation process for training had little impact. However, as shown in Table \ref{tab:my-table5}, when noise documents were provided alongside evidential documents, this process improved the noise robustness of the compressor.

\chapter{Conclusion}\label{chap:conclusion}
In this paper, we introduce a simple training approach, ACoRN, aimed at improving noise robustness during the compression of retrieved documents. We improved the compressor by focusing on two key aspects to ensure practical usability. First, to be applicable in real-world scenarios, the compressor must be noise-robust not only against irrelevant or insufficient documents but also against factual error documents that provide incorrect information. Second, to address positional bias in language models, we heuristically defined evidential documents containing the answer string. These documents enabled optimal summarization of key information while mitigating positional bias through their distribution across various positions. Additionally, we establish two benchmarks to demonstrate the noise robustness of the compressor. The first benchmark measures its ability to preserve the answer string ratio and compression ratio during summarization. The second benchmark assesses performance changes when noisy documents are incorporated into evidential documents. We verify that ACoRN notably increases the preserving answer string ratio and improves noise robustness.

\newpage
\renewcommand\bibname{Bibliography}










\end{document}